%% file: longctx.tex
\pdfoutput=1

\documentclass[11pt]{article}

\usepackage[preprint]{acl}

\usepackage{times}
\usepackage{latexsym}
\usepackage{inconsolata}

\usepackage[utf8]{inputenc} 
\usepackage[T1]{fontenc}    
\usepackage{hyperref}       
\usepackage{url}            
\usepackage{booktabs}       
\usepackage{amsfonts}       
\usepackage{nicefrac}       
\usepackage{microtype}      

\usepackage{graphicx} 
\usepackage{svg}
\usepackage{wrapfig}
\usepackage{makecell}
\usepackage{multirow}
\usepackage{colortbl}
\usepackage{algorithm}
\usepackage[noend]{algpseudocode}
\usepackage{amsmath}
\usepackage{subcaption}
\usepackage{tablefootnote}

\newcommand{\modelname}{Qwen2-UtK}

\title{Untie the Knots: An Efficient Data Augmentation Strategy for Long-Context Pre-Training in Language Models}

\author{%
 \textbf{Junfeng Tian\textsuperscript{1*}},
 \textbf{Da Zheng\textsuperscript{1*}},
 \textbf{Yang Chen\textsuperscript{2}}, 
 \textbf{Rui Wang\textsuperscript{3}}, \\
 \textbf{Colin Zhang\textsuperscript{1}},
 \textbf{Debing Zhang\textsuperscript{1\dag}} \\
 \textsuperscript{1} Xiaohongshu Inc, \qquad \textsuperscript{3} Decilion, \\
 \textsuperscript{2} School of Computer Science and Technology, East China Normal University\\
 \texttt{\{tianjunfeng, zhengda, martin, dengyang\}@xiaohongshu.com}\\
 \texttt{yangchen@stu.ecnu.edu.cn} \quad \texttt{mars.wang@disiling.cn} \\
}

\begin{document}

\maketitle
\renewcommand{\thefootnote}{\fnsymbol{footnote}} 
\footnotetext[1]{Equal contributions.} 
\footnotetext[2]{Corresponding author.} 

\begin{abstract}
Large language models (LLM) have prioritized expanding the context window from which models can incorporate more information. However, training models to handle long contexts presents significant challenges. These include the scarcity of high-quality natural long-context data, the potential for performance degradation on short-context tasks, and the reduced training efficiency associated with attention mechanisms. In this paper, we introduce Untie the Knots (\textbf{UtK}), a novel data augmentation strategy employed during the continue pre-training phase, designed to efficiently enable LLMs to gain long-context capabilities without the need to modify the existing data mixture. In particular, we chunk the documents, shuffle the chunks, and create a complex and knotted structure of long texts; LLMs are then trained to untie these knots and identify relevant segments within seemingly chaotic token sequences. This approach greatly improves the model's performance by accurately attending to relevant information in long context and the training efficiency is also largely increased. We conduct extensive experiments on models with 7B and 72B parameters, trained on 20 billion tokens, demonstrating that UtK achieves 75\% and 84.5\% accurracy on RULER at 128K context length, significantly outperforming other long context strategies. The trained models will open-source for further research.
\end{abstract}

\input{chapter/1_intro}

\input{chapter/2_related_work}

\input{chapter/3_method}

\input{chapter/4_experiment}

\input{chapter/5_conclusion}


\medskip


\bibliography{custom}

\newpage
\appendix

\input{chapter/appendix}

\end{document}

%% file: chapter/1_intro.tex
\section{Introduction}


\begin{figure}[t]
    \includegraphics[width=\columnwidth]{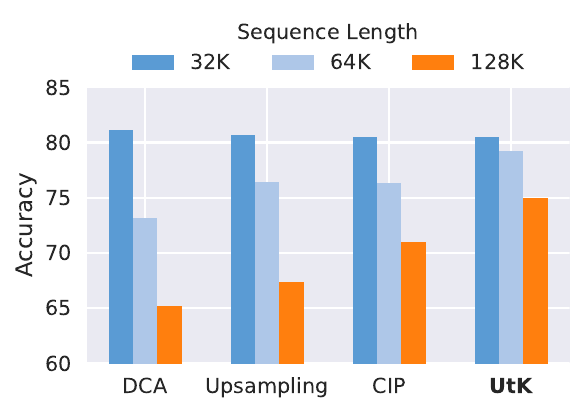}
    \caption{Comparison of various long-context strategies based on the Qwen2-base (7B) model on the RULER benchmark. UtK more effectively maintains performance at the 128K context length.}
    \label{fig:intro}
\end{figure}


For the past few years, large language models (LLM) research has prioritized expanding the context window from which models can incorporate more information \cite{gpt3, anthropic2023, gpt4, gemini15}. This emphasis stems from the recognition that a wider context window allows models to incorporate a larger amount of new, task-specific information not found in the training data at inference time, leading to improved performance in various natural language tasks \cite{caciularu2023peekacrossimprovingmultidocument,bairi2023codeplanrepositorylevelcodingusing,mazumder2024lifelongcontinuallearningdialogue,jiang2024longragenhancingretrievalaugmentedgeneration,gur2024realworldwebagentplanninglong}.

However, training transformer-based \cite{transformer} models to handle long contexts effectively poses significant challenges due to the lower training efficiency and the quadratic computational cost of attention mechanisms in long-context models. As a result, many approaches treat long-context extension as a distinct stage. Training-free methods for length extropolation, such as those that modify Rotary Position Embedding (RoPE)\cite{rope}, often fail to deliver satisfactory performance. Continue pre-training approaches \citep{llama31, ChatGLMLong2024, gunter2024afm} aimed at improving long-context performance encounter a critical issue: the scarcity of sufficiently long training texts, which introduces a bias in training data. Texts ranging from 32K to 128K tokens are rare and typically consist of books and code. To mitigate this, methods like LLama3.1 and GLM-Long use per source upsampling and artificial long texts (e.g., concatenated similar documents) to increase the presence of long sequences in the training data. However, these approaches alter the data distribution, making it challenging to achieve a model that performs well on both long-context and short-context tasks while maintaining efficiency.

In this paper, we introduce a novel augmented training strategy called \textbf{Untie the Knots} (UtK), designed to enhance the long-context capabilities of LLMs without altering the existing data mixture. UtK employs an augmentation recipe that helps the model to adapt to longer input sequences more effectively. Specifically, this strategy involves chunking, shuffling, and reconstructing the input documents, encouraging the model to learn to attend to relevant segments of the same documents while skipping unrelated segments in between. Furthermore, we introduce a backtracing task for the model to explicitly locate all the corresponding segments in the correct order, which largely improves the accuracy of finding the original context in longer ranges. 
This strategy, illustratrated in Figure~\ref{fig:utk_main}, ensures that the model maintains a coherent understanding between and beyond documents, enhancing its ability to handle short and long contexts at the same time.

To assess the effectiveness of Untie the Knots, we conducted continue pre-training of language models with 7B and 72B parameters on 20 billion tokens. Our results demonstrate that UtK outperforms the ABF baseline and other data strategies, such as upsampling, as shown in Figure~\ref{fig:intro}. It also significantly exceeds the performance of training-free extropolation methods like YaRN \cite{yarn} and Dual Chunk Attention (DCA) \cite{an2024dca}. Specifically, our models show significant improvements on widely-used benchmarks, achieving 15.0\% increase in performance on RULER and 17.2\% increase on LV-Eval for 128K tasks, which are both over 90\% of the performance on 32K contexts. We will open-source the Qwen2-7B-UtK-128k and Qwen2-72B-UtK-128k base models to facilitate further research in this area.

Our contributions are as follows:
\begin{enumerate}
\item We introduce Untie the Knots (UtK), an innovative data augmentation strategy designed to improve the long-context capabilities of large language models. This method enhances both training efficiency and model performance on long-context tasks.
\item We conduct extensive experiments on 7B and 72B models, trained on up to 20 billion tokens. Our results demonstrate that UtK significantly outperforms existing data strategies, such as length upsampling and DCA, across multiple benchmarks.
\item We will open source two well-trained models, \modelname-7B-base 128K and \modelname-72B-base 128K, to facilitate further research and development of the field of long-context language models.
\end{enumerate}

%% file: chapter/2_related_work.tex
\section{Related Work}

\begin{figure*}[t!]
    \centering
    \includegraphics[width=\textwidth]{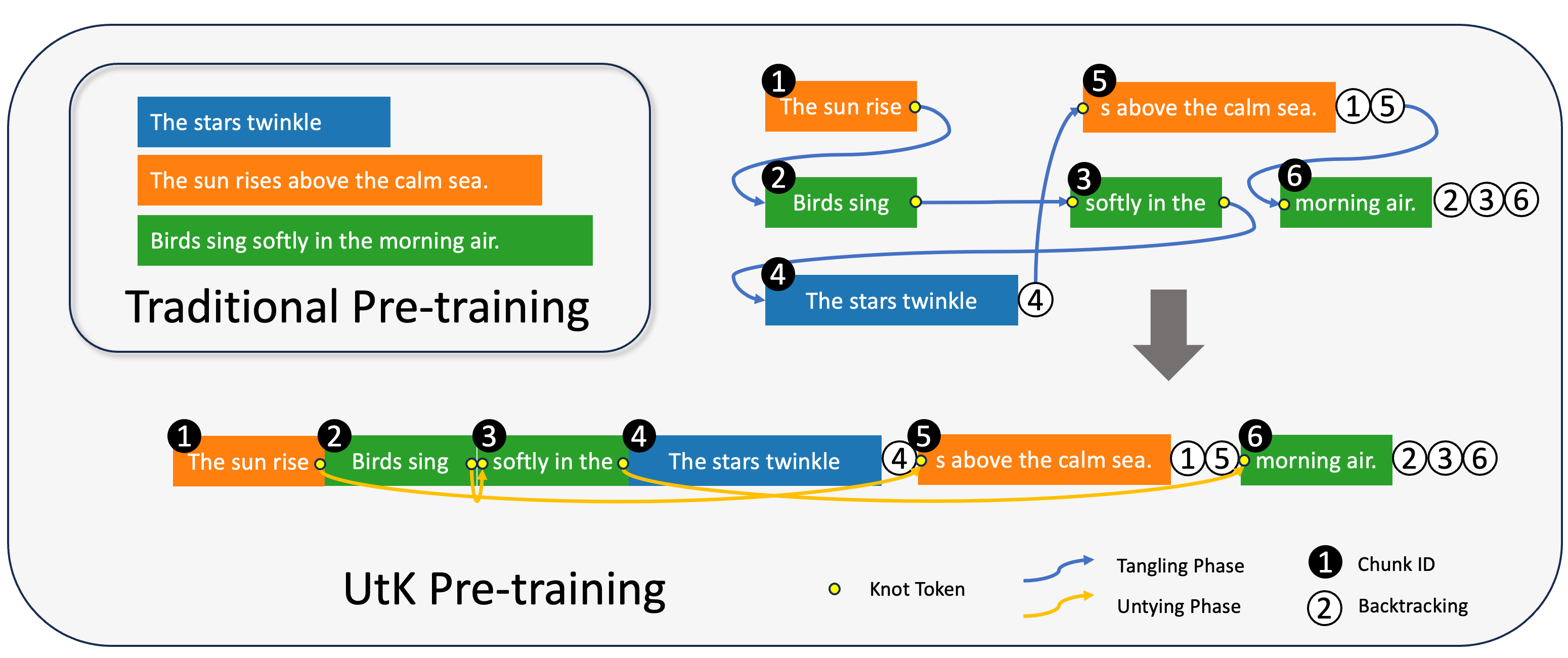}
    \caption{Illustration of the \textbf{UtK} Pre-training process. In the \textbf{Tangling phase}, documents are split into chunks, which are then randomly tied together. \textbf{Knot Tokens} are inserted at the split points to guide the model in locating the partitions during the \textbf{Untying phase}. The \textbf{Chunk IDs} of each chunk are appended to the last chunk of the document to help the model learn to correctly \textbf{backtrace} the original document structure.}
    \label{fig:utk_main}
\end{figure*}

\subsection{Long Context in LLMs}




Long Document continue pre-training has become a crucial step in enhancing long-context capabilities in foundational models. Plenty of leading foundational models \citep{gemini15, llama31, yang2024qwen2, ChatGLMLong2024, gunter2024afm} have emphasized the importance of RoPE’s positional encoding and the upsampling of lengthy data. For example, models like LLaMA 3.1 \citep{llama31} and Phi-3 \citep{abdin2024phi3} leverage the Long RoPE method \citep{ding2024longrope} to extend their context windows, while Qwen2 \citep{yang2024qwen2} utilizes the YARN and Dual Chunk Attention mechanisms \citep{yarn, an2024dca} to increase the context length to 128k. Additionally, GLM Long \citep{ChatGLMLong2024} and Apple’s AFM \citep{gunter2024afm} scale the RoPE base frequency \citep{men2024base} to improve generalization across varying sequence lengths.

One series of works manipulate the order of training tokens to achieve similar goals. For instance, UL2 \citep{tay2022ul2} designs mixture of denoisers (MoD) objective to adapt the model to different tasks. FIM \citep{bavarian2022efficienttraininglanguagemodels} applied data transformation by splitting documents into three random segments and rearranging them with sentinel tokens. FIM gives model ability to generate content conditioned on both prefix and suffix, which is essential on tasks like code editing. In-context Pretraining \citep{shi2024iclm} proposed to train on a sequence of related documents to explicitly encourage the model to read and reason across document boundaries.

Another series of works, such as PoSE \citep{zhu_pose_2024}, introduce large random gaps within the same document to help the model become familiar with once OOD relative distances. LongSkywork \citep{zhao2024longskywork} proposed Chunk Interleaved pre-training where documents are split into segments, which are then arranged in an interleaved fashion to form pseudo long-context pre-training samples. Our approach differs by employing a multi-hop strategy that involves splitting, shuffling, and merging data, thereby enhancing long-context capabilities through a more straightforward yet effective training process.

\subsection{Rotary Position Embedding}
Rotary Position Embeddings (RoPE) \cite{rope} have effectively encoded positional information in transformer-based models, yet they struggle to generalize beyond their trained sequence lengths. To overcome this limitation, various methods have been proposed to extend RoPE’s context window for handling longer sequences.

Position Interpolation (PI) \cite{position_interpolation} extends RoPE by linearly interpolating the position index within the original context window. While effective, PI’s uniform frequency scaling may limit the model’s ability to capture high-frequency features. NTK-Aware and NTK-By-Parts \cite{ntk} introduce nonlinear strategies to address PI’s limitations. NTK-Aware adjusts RoPE’s base frequency, while NTK-By-Parts selectively scales different frequency components to better preserve local token relationships, enhancing the model’s capacity to manage longer sequences. YaRN \cite{yarn} builds on NTK-By-Parts by introducing a temperature to scale attention logits before softmax, further improving language modeling performance on long-context tasks. Adjusted Base Frequency (ABF) \cite{llamalong} modifies RoPE’s base frequency to 50,000, empirically demonstrating lower perplexity and extended context capabilities. In this work, we adopt the ABF method, adjusting RoPE’s base frequency according to \citet{men2024base}, who suggest that the base of RoPE sets a context length boundary, providing a minimum base value necessary for achieving specific context lengths.

%% file: chapter/3_method.tex
\section{Method}








\begin{figure}[t]
    \centering
    \includegraphics[width=0.8\columnwidth]{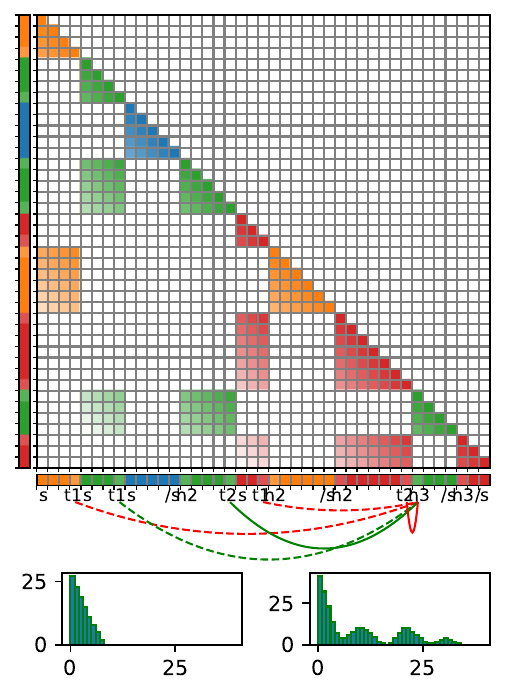}
    \caption{The top panel shows the UtK-augmented expected conditional information for the same four documents, while the bottom panel displays the changes in the histogram of relative positional embedding distances from the original to the UtK-augmented.}
    \label{fig:attention}
\end{figure}

In this section, we describe our Untie the Knots (UtK) augmentation strategy, aiming for effectively enhancing language models' long context abilities. The illustration of the method is shown in Figure \ref{fig:utk_main}. See \autoref{sec:utk_algorithm} for more details.


\subsection{Tangling Phase}

\paragraph{Chunking} First, we chunk documents within target sequence length into several chunks, the split points are randomly chosen. Knot tokens are added before and after the split point. Chunk ID is prepended at each chunk

\paragraph{Tying} Chunks are shuffled and randomly tied together. Considering the order of chunks of the same document may affect the result. We tried two strategies, preserve order and no preserver order.

\subsection{Backtracing}

After the final chunk of each document, we inserted the chunk IDs of this document, the model will enchant the ability to do backtracing in long range. A sentinel token is included to trigger the backtracing output. We masked the loss on both the knot tokens and sentinel token to keep the model from generating them.

\subsection{Untying Phase}
UtK turns the irrelevant documents into a complex myth. when the language model meet the ``head knot'', it's prompted to search its context for the unique matching "tail knot." Only upon finding this correct match can the model reconstruct the fragmented document and proceed with its usual language processing. Furthermore, if a document is split into multiple chunks, the model must successfully identify and connect all related knots to fully restore the original context.

\subsection{Longer than claimed}
As indicated in the histogram in Figure \ref{fig:attention}, we realized that even when UtK is enabled, distances which is near the training sequence length are still rare in training data. So we purpose to use slightly longer sequence length than claimed max sequence length in training to get better performance.




%% file: chapter/4_experiment.tex
\section{Experimental Setting}


\subsection{Training Data}

\begin{figure}[t]
    \centering
    \includegraphics[width=0.8\columnwidth]{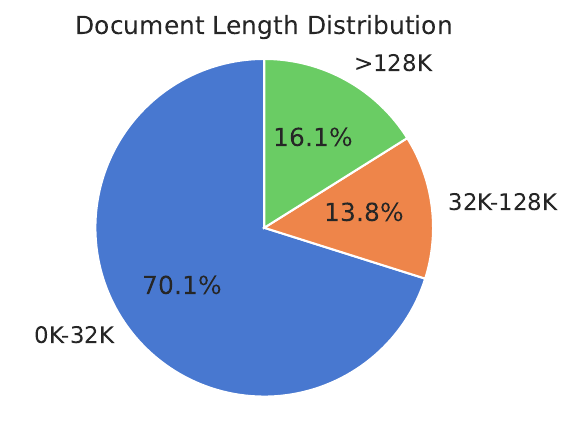}
    \caption{Distribution of document lengths categorized by token counts. The ratios represent the number of tokens within each document length category proportional to the total number of tokens.}
    \label{fig:length}
\end{figure}

Following \citet{llama, llama2, llama31, yang2024qwen2}, who emphasize the influence of data quality and diversity in training models, our curated dataset incorporates sources such as Common Crawl, books, Wikipedia, code, and academic papers. Additionally, our dataset is multilingual, with a significant portion of the data in English and Chinese. For continued training, we employ a quality classifier to filter high-quality data. After filtering, we randomly sample a total of 300 billion tokens for pre-training. \autoref{fig:length} illustrates the distribution of document lengths, where 70\% of the data falls within the 0-32K token range.

\subsection{Model Details}

We continued pre-training the Qwen2 models with a sequence length of 128K tokens, as they were initially trained on sequences of only 32K tokens. The AdamW optimizer \cite{adamw} is used for optimization, with parameters $\beta_1 = 0.9$ and $\beta_2 = 0.95$, alongside a cosine learning rate schedule starting at 1e-5 and decaying to 1e-6, with 200 warmup steps. Due to the models’ long context windows, ring attention (cp=4) \cite{liu2023ring} and flash attention \cite{flashattn} are employed to reduce memory consumption. The training setup involves 128 H800 GPUs across 16 nodes, with a batch size of 4 million tokens. Training the 7B parameter models on 20B tokens takes 15 hours, while the 72B models require 5.5 days to complete training on the same amount of data.
For each document, with a certain probability $p$, we split it into $n$ parts: Chunk$_1$, Chunk$_2$, $\cdots$, Chunk$_n$. This split occurs after tokenization, making it an on-the-fly solution that can be applied to other architectures (e.g., Mamba\cite{gu2024mamba}). The split position is performed uniformly at random. We conduct experiments using two probabilities, 30\% and 80\%, representing low and high splitting rates by default.


\subsection{Comparison Methods} \label{sec:comparison_methods} 

We compare UtK against the following methods:

\paragraph{CT} In the naive continued pre-training experiment, we increased the training sequence length to 128K and trained on 20 billion tokens. Since the models were already pre-trained on 128K data, DCA was not applied during the inference stage.

\paragraph{ABF} We increased the base frequency $b$ of RoPE \citep{llamalong} from $1e6$ to $5e6$, which is approximately the recommended base frequency as proposed by \citet{men2024base}. Note that the $5e6$ base frequency was used in all experiments except for the naive CT baseline in this paper.

\paragraph{Upsampling} Following  \citet{fu2024dataengineeringscalinglanguage}, we applied per-source length upsampling to maintain a fixed domain mixture ratio. Documents longer than 32K tokens were upsampled by a factor of five, without altering the overall domain mixture ratio.

\paragraph{AttnMask} As suggested by \citet{llama31}, an inter-document attention mask is essential during continued pre-training for long context. We applied this strategy in our experiment. Note that this strategy cannot be combined with UtK, as UtK requires the model to have full attention to locate the corresponding knots.

\paragraph{Synthetic} \citet{xiong2024artificialneedlesrealhaystacks} demonstrated that fine-tuning LLMs using specially designed synthetic data can significantly enhance long-context understanding. Inspired by their approach, we constructed five types of synthetic datasets focused on specific tasks: sorting, multi-hop reasoning, state tracking, similarity retrieval, and attribute inclusion. Each dataset had a context length of 128K tokens. In this experiment, 30\% of the original data mixture was replaced with synthetic data.

\paragraph{CIP} Following the optimal CIP-2 configuration from \citet{zhao2024longskywork}, each document was randomly split into two chunks, which were then interleaved in a pattern such as $D_1^1, D_2^1, D_3^1, D_1^2, D_2^2, D_3^2$.

\section{Results}
\subsection{Main Results}

\subsubsection{Long Tasks}

\paragraph{Datasets \& Metrics}

To quantify the long context adaptation rate, we mainly focus on evaluate long-context language models on test sets with configurable sequence length. We use two widely recognized benchmarks: RULER \cite{hsieh2024ruler} and LV-Eval \cite{yuan2024lveval}. RULER generates synthetic examples to assess long-context capabilities beyond simple in-context recall, comprising 13 tasks across 4 categories (i.e, NIAH, VT, CWE+FWE, and QA). We use the base model prompt template and report the average score across these 13 tasks.

LV-Eval consists of two main tasks, single-hop QA and multi-hop QA, across 11 bilingual datasets. To reduce the influences from prompt engineering and minimize bias in automated evaluations, we assess 3-shot performance at 32K and 128K context lengths. We exclude the factrecall-en and factrecall-zh datasets, as factrecall-en and factrecall-zh are designed for pressure test of ``needle in haystack'', and they are not relevant to our work. Instead, we focus on evaluating real-world language tasks. We report the average F1 score or ROUGE score across the remaining 9 datasets. For all tasks except dureader-mixup and cmrc-mixup, we use a keyword-recall-based F1 metric, utilizing annotated answer keywords and a word blacklist. For cmrc-mixup, we apply the F1 metric with a word blacklist, and for dureader-mixup, we use the ROUGE-L metric with a word blacklist.



\paragraph{Results.} 
The results in \autoref{tab:ruler} and \autoref{tab:lveval} highlight our model's effectiveness across various long-context evaluation benchmarks. Detailed values for different datasets are provided in  \autoref{app:sec:results}. On the RULER benchmark, our model, \modelname-base (7B), consistently outperforms most other models at the 128K context length, achieving an average score of 75.0 —significantly higher than Qwen2-base by 15.0\% and Llama3.1-base by 13.5\%. This demonstrates that \modelname-base is particularly robust in handling extended contexts, maintaining strong performance as context length increases. 
We also applied UtK to Llama3.1-base, a model with a 128K context length, to evaluate its robustness. Llama3.1-UtK-base was trained using llama-rope, and UtK demonstrated an improvement of 11.6\% in performance.

In the LV-Eval benchmark, which emphasizes real-world language tasks, our model once again exhibits superior performance. We consider 32K as a performance upper bound for 128K, and in comparing different models, we find that our model consistently performs well on both single-hop and multi-hop QA tasks at 128K, further indicating its exceptional robustness.

\begin{table}[t]
\normalsize
\setlength{\tabcolsep}{1.8pt}
\begin{tabular}{lcccc}
\toprule
Models & 32K & 64K & 128K & Rate\\
\midrule
\textsc{API Model} \\
Gemini-1.5-pro$^\S$ & \textbf{95.9} & \textbf{95.9} & \textbf{94.4} & \textbf{98\%} \\
GPT-4-1106-preview$^\S$ & 93.2 & 87.0 & 81.2 & 87\% \\
\midrule
\textsc{Instruct Model} \\
Mistral (7B)$^\S$ & 75.4 & 49.0 & 13.8 & 18\% \\
LWM (7B)$^\S$ & 69.1 & 68.1 & 65.0 & 94\% \\
Llama3.1 (8B)$^\S$ & 87.4 & 84.7 & 77.0 & 88\% \\
\midrule
\textsc{Base Model} \\
Mistral-base (7B)$^\S$ & 77.2 & 52.3 & 8.0 & 10\% \\
LWM-base (7B)$^\S$ & 64.6 & 61.3 & 59.0 & 91\% \\
Qwen2-base (7B)$^\ddag$ &  81.1 &  73.2  & 65.2 & 80\% \\
Llama3.1-base (8B)$^\dag$ & 90.2 & 80.4 & 66.1 & 73\% \\
\midrule
\textsc{Long-Context} \\
Qwen2-CT (7B) & 78.2 & 73.6 & 54.8 & 70\% \\  
Qwen2-ABF (7B) & 78.9 & 75.2 & 65.9 & 84\%\\
Qwen2-Upsampling (7B) & 80.7 & 76.4 &  67.4 & 84\%  \\
Qwen2-AttnMask (7B) & 80.4 & 75.6  & 72.0 & 90\% \\
Qwen2-Synthetic (7B) & 83.2 & 80.5 & 72.7 & 87\% \\
Qwen2-CIP (7B)& 80.5 & 76.3 & 71.0 & 88\% \\ 
\rowcolor{gray!30} \modelname-base (7B) & 80.5 & 79.2 & 75.0 & 93\% \\
\rowcolor{gray!30} Llama3.1-UtK-base (8B)$^\dag$ & 88.8 & 83.6 & 73.8 & 83\% \\
\midrule
\multicolumn{4}{l}{\textsc{70B Model}} \\
Llama3.1 (70B)$^\S$ & \underline{94.8} & 88.4 & 66.6 & 70\% \\
Qwen2 (72B)$^\S$ & 94.1 & 79.8 & 53.7 & 57\% \\
Llama3.1-base (70B)$^\dag$ & 91.7 & 84.6 & 66.0 & 72\% \\
Qwen2-base (72B)$^\ddag$ &  93.3  & 85.9 & 78.0 & 84\% \\
\rowcolor{gray!30} \modelname-base (72B) & 93.3 & \underline{90.6} & \underline{84.5} & \underline{94\%}  \\
\bottomrule
\end{tabular}
\caption{Performance on the RULER benchmark. $^\dag$Llama3.1-base was inferred with vllm. $^\ddag$For Qwen2-base (7B, 72B), we used vLLM \href{https://github.com/vllm-project/vllm/pull/6139}{DCA branch} for tasks over 32K tokens as suggested by Qwen Team. $^\S$ results are sourced from RULER.}
\label{tab:ruler}
\end{table}

\begin{table*}[t]
\centering
\setlength{\tabcolsep}{3.2pt}
\begin{tabular}{lccccccc}
\toprule
\multirow{2}{*}{Models}  & \multicolumn{3}{c}{32K}  & \multicolumn{3}{c}{128K}  & \multirow{2}{*}{Rate} \\
 & Average & {Single-hop} & {Multi-hop} & Average & {Single-hop} & {Multi-hop} \\
 \midrule
Llama3.1-base (8B) & 29.16 & 43.18 & 17.94 & 23.90 & 33.49 & 16.23 & 82.0\% \\
Qwen2-base (7B) & \textbf{29.88} & 42.72 & 19.61 &  23.94 & 35.41 & 14.76 & 80.1\% \\
\rowcolor{gray!30} Llama3.1-UtK-base (7B) & 29.63 & \textbf{43.36} & 18.65 & 26.89 & 37.85 & 18.13 &  90.8\% \\
\rowcolor{gray!30} \modelname-base (7B) & 29.36 & 39.79 & \textbf{21.02} & \textbf{28.06} & \textbf{38.99} & \textbf{19.32} & \textbf{95.6\%} \\
\midrule
Llama3.1-base (70B) & 30.38 & 42.19 & 20.94 & 23.07 & 31.02 & 16.71 & 76.0\% \\
Qwen2-base (72B) & \textbf{32.37} & \textbf{44.22} & 22.88 & 27.40 & 37.38 & 19.42 & 84.7\% \\
\rowcolor{gray!30} \modelname-base (72B) & 32.24 & 43.54 & \textbf{23.20} & \textbf{32.10} & \textbf{43.92} & \textbf{22.65} & \bf 99.6\% \\
\bottomrule
\end{tabular}
\caption{Performance on LV-Eval benchmark.}
\label{tab:lveval}
\end{table*}

\subsubsection{Short Tasks}

\paragraph{Datasets \& Metrics}  Previous research \cite{llamalong, llama31} has identified a model performance tradeoff between short and long tasks. To evaluate our models' performance on short tasks, we conducted tests on a series of widely recognized benchmarks. Specifically, we assess our models using three categories of datasets: Understanding, Code, and Math. For Understanding, we assess 5-shot performance on Natural Questions \cite{natural_questions}  and TrivialQA \cite{triviaqa}, and 3-shot Chain-of-Thought performance on BIG-Bench Hard \cite{bbh}. In the Code category, we measure pass@1 on HumanEval \cite{humaneval} and 3-shot performance on the sanitized MPBB benchmark \cite{mbpp}. For Math, we evaluat top-1 accuracy on the 4-shot GSM8K dataset \cite{gsm8k}. These metrics provide a comprehensive assessment of the models' capabilities across diverse tasks.

\paragraph{Results.} \autoref{tab:short_task} presents the average scores across different model sizes. First, we analyze the impact of data on the model performance and find that using our data achieves comparable performance to the base model, with a slight decrease. Second, after removing the impact of the data, we observe that our method's metrics are similar to those of the CT baseline. These findings suggest that our methods enable language models on long-context tasks while maintaining performance on short tasks.

\begin{table*}[]
\centering
\setlength{\tabcolsep}{2pt}
\begin{tabular}{lccccccc|ccc}
\toprule
\multirow{2}{*}{Model} & \multicolumn{3}{c}{Understanding} & \multicolumn{2}{c}{Code} & Math & \multirow{2}{*}{Avg.} & \multicolumn{3}{c}{RULER}  \\
\cmidrule(r){2-7} \cmidrule(l){9-11}
 & \makecell{BBH\\3shot}  & \makecell{NQ\\5shot} & 
 \makecell{TriviaQA\\5shot} & \makecell{HumanEval\\0shot} & \makecell{MBPP\\3shot} & \makecell{GSM8K\\4shot} &  & 4K & 8K & 16K \\
\midrule
Llama3.1-base (8B) & 63.9  & 33.5 & 80.2 & 35.4 & 54.5 & 58.0 & 54.2 & 94.3 & 92.1 & 92.3\\
\rowcolor{gray!30} Llama3.1-UtK-base (8B) & 61.9  & 34.0 & 79.6 & 38.4 & 54.6 & 59.3 & 54.6 & 94.6 & 92.2 & 91.7 \\
Qwen2-base (7B) & 61.4 & 30.3 & 70.2 & 46.3 & 64.6 & 80.9 & 59.0 & 90.6 & 85.0 & 82.3 \\
\rowcolor{gray!30} Qwen2-CT-base (7B) & 61.1  &  29.6 &  70.3 & 44.5 & 66.2 &  77.6 & 58.1 & 92.8 & 85.8 & 83.1 \\
\rowcolor{gray!30} \modelname-base (7B) & 61.6  & 29.5 & 70.2 & 45.1 & 64.2 & 78.1 & 58.2 & 90.6 & 85.0 & 82.0 \\
\midrule
Llama3.1-base (70B) & \bf 81.0 & \bf 49.3 & \bf 91.2 & 59.2 & 72.8 & 82.3 & 72.6 & 95.8 & 94.5 & 93.0 \\
Qwen2-base (72B)   & 79.8 & 45.6 & 88.0 & \bf 61.6 & \bf 76.9 & \bf 88.8 & \bf 73.3 & \bf 97.1 & \bf 95.6 & 94.3 \\
\rowcolor{gray!30} \modelname-base (72B) & 80.6 & 45.0 & 87.6 & 61.0 & 75.9 &  87.8  & 73.0 & 95.0 & 93.8 & \bf 94.7 \\
\bottomrule
\end{tabular}
\caption{Performance on standard short-context benchmarks.}
\label{tab:short_task}
\end{table*}

\subsection{Ablation Analysis}

\begin{figure*}[t]
    \centering
    \includegraphics[width=1.0\textwidth]{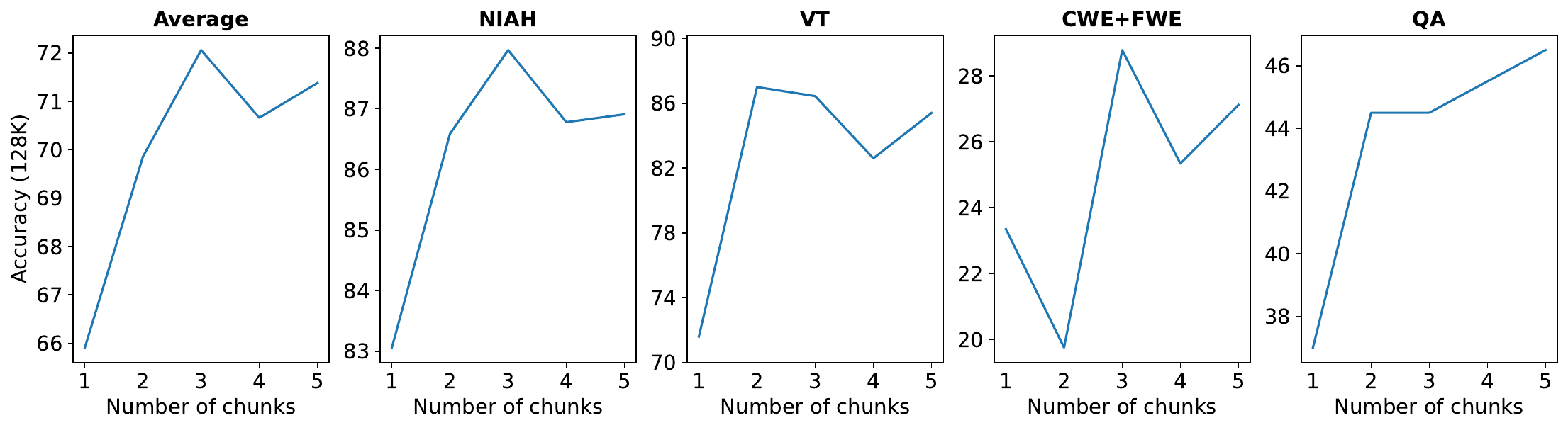}
    \caption{Performance with varying numbers of chunks on the RULER 128K benchmark.}
    \label{fig:cut}
\end{figure*}

\begin{table}[h]
\setlength{\tabcolsep}{0.8pt}
\begin{tabular}{lccccccc}
\toprule
Chunk & Average & NIAH & VT & \makecell{CWE+\\FWE} & QA \\
\midrule
\modelname-base & 75.0 & 90.3 & 97.6 & 29.9 & 48.0 \\
- Disrupt order & 73.0 & 90.4 & 97.8 & 17.4 & 46.0 \\
- W/o backtracing & 74.3 & 91.3 & 94.8 & 23.5 & 46.5 \\
UtK (30\%) & 73.1 & 88.8 & 94.8 & 28.5 & 44.0 \\
- Disrupt order & 72.3 & 89.0 & 89.8 & 21.7 & 47.0 \\
- W/o backtracing & 70.8 & 88.5 & 86.2 & 21.2 & 42.0 \\
\bottomrule
\end{tabular}
\caption{Ablation Study. UtK (30\%) denotes applying UtK to 30\% of the documents. Disrupt order indicates that the sequential order of the chunks within the documents is not preserved. W/o backtracing signifies that backtracing is not applied during the process.} 
\label{tab:ablation}
\end{table}

We have conducted ablation analyses on two key design choices in the training strategy: (1) the optimal number of chunks for long-context training, and (2) the effects of each designed component. We have performed the ablation study on 7B models with 20B training tokens and evaluated them with the RULER benchmark. The results are illustrated in \autoref{fig:cut} and \autoref{tab:ablation}.

\paragraph{Number of Chunks}  When evaluating the number of chunks, we find that using 2 or 3 chunks yields the best performance on the NIAH, VT, and CWE+FWE datasets. For the QA dataset, we observe that increasing the number of chunks improves the model's reasoning ability, suggesting that more complex training benefits QA tasks. We have also experimented with combining these approaches, which resulted in even better performance.
We tried dividing the text into chunks of 1K tokens each, which resulted in an average score of 68.92. This indicates that a higher number of chunks can increase task complexity, potentially hindering the model's learning process.

\paragraph{Training Strategy}  In comparing different training strategies, we observe that maintaining partial order and incorporating the tracing task are both essential for long-context learning. We reckon that keeping the partial order encourages the model to attend to longer but related chunks, while the tracing task requires the model to provide the "correct" untie solution, as later segments cannot typically correct errors in earlier ones. Finally, we find that a higher probability of UtK is also necessary to improve training efficiency.

\subsection{Training Efficiency}

\begin{figure}[t]
    \centering
    \includegraphics[width=0.9\columnwidth]{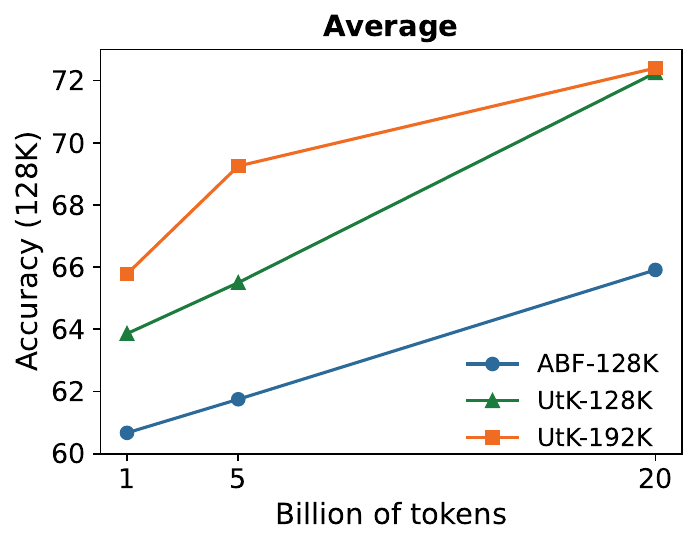}
    \caption{Training Efficiency}
    \label{fig:tokens}
\end{figure}

As illustrated in Figure~\ref{fig:tokens}, we compare the baseline and UtK training methods by progressively increasing the number of training tokens to determine the required amount for effective long-context extension. We also include experiments with a longer sequence length of 192K to assess whether even longer context would enhance performance when still evaluated on the 128K tasks.

Our findings indicate that: 1) Our approach UtK does have a higher training efficiency compared with the baseline regardless of how many training tokens are used, and the performance gains are steady. 2) Training on a 192K sequence length does increase the training efficiency at both the 1B and 5B token levels but the grains are diminishing when we reach 20B tokens. 3) Most significantly, with only 1B tokens, UtK-192K can already reach ABF's performance after 20B tokens training.

\subsection{Attention Visualization}

To visually represent the changes in attention of the model trained with UtK at a length of 128k, we have plotted the attention maps for the model trained on 128k lengths. We compared the original Qwen2 model, the ABF baseline, and the model trained with UtK on text lengths of 128k. Although the ABF-trained baseline can already accurately locate information within the same document, the model trained with UtK exhibits more attention on long-range dependencies within the same document, thereby reducing the loss of long-range relevant information. The detailed content and explanations of the plots can be found in the appendix.

\begin{figure}[t]
    \centering
    \includegraphics[width=0.92\columnwidth]{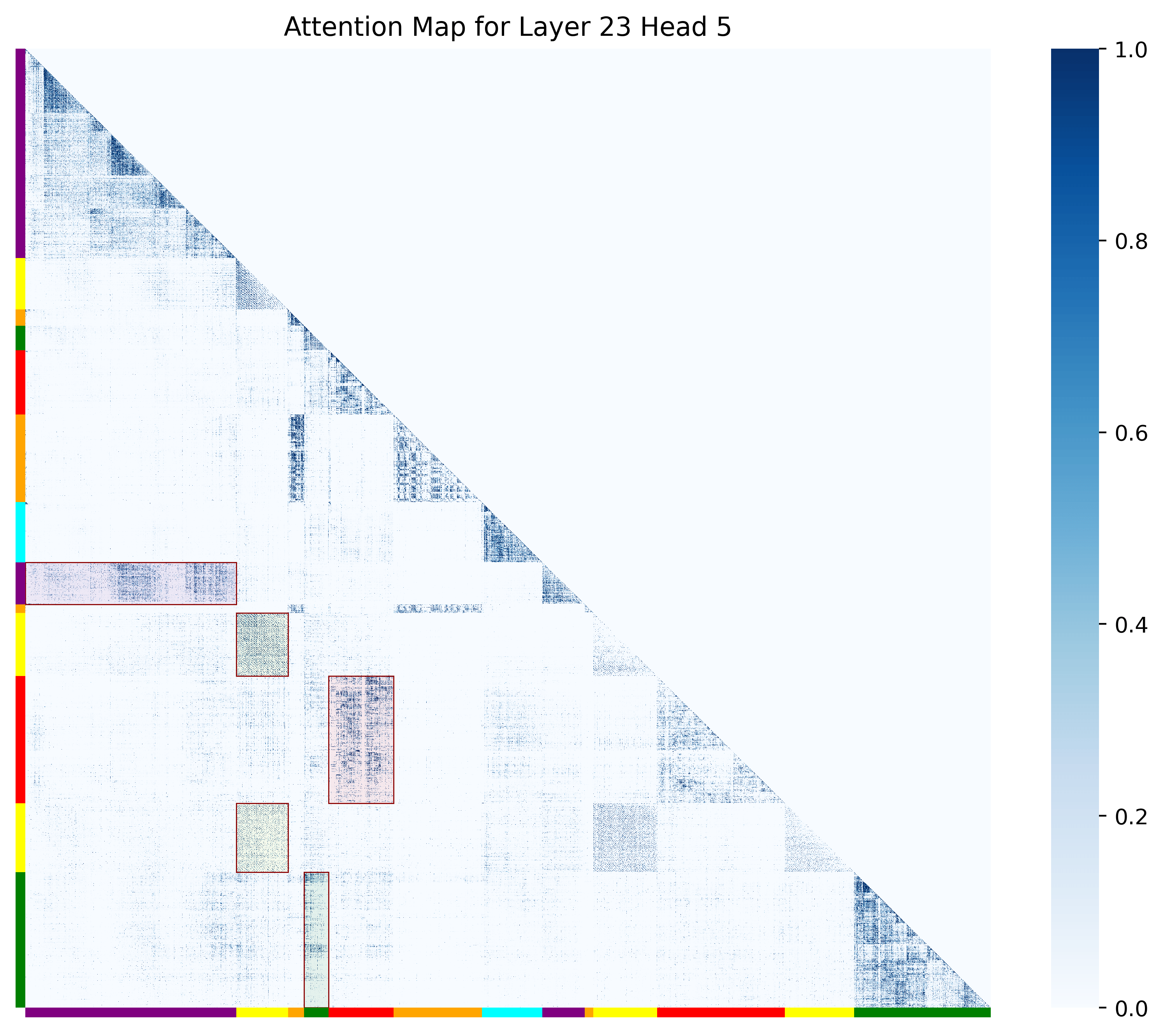}
    \caption{\modelname-base 7B}
    \label{fig:attn2}
\end{figure}

%% file: chapter/5_conclusion.tex
\section{Conclusion}




In this paper, we proposed UtK, an augmentation recipe to adapt the model to longer context more efficiently and more effectively. We have trained and open sourced Qwen2-7B-UtK-128k and Qwen2-72B-UtK-128k base models with superior performance to the base models, as well as other long context enhancement strategies including upsampling and DCA. We have also introduced a long context adaption rate as an evaluation metric to measure how well such models have been adapted to the long context tasks.
In addition to the performance gain, our method also demonstrates a large increase of training efficiency.
We have open sourced the two models and sincerely hope to see our approach applied to more datasets and model training in the community.

\section{Limitations}

Although being efficient among continue training methods, due to the limitation of training tokens and practice patterns. As a result, it can only perform adaptation or transfer learning based on the model's original ability. Acquiring new abilities, such as solving complex problems within long context, is not feasible and may require further specialized training. Our experiments are also limited to the datasets we use. Our method applied to other datasets of different languages or genres might lead to different results.

%% file: chapter/appendix.tex



\section{Attention Visualization at 128k Lengths}

\begin{figure*}[h!]
    \centering
    \begin{subfigure}[b]{0.49\textwidth}
        \centering
        \includegraphics[width=\textwidth]{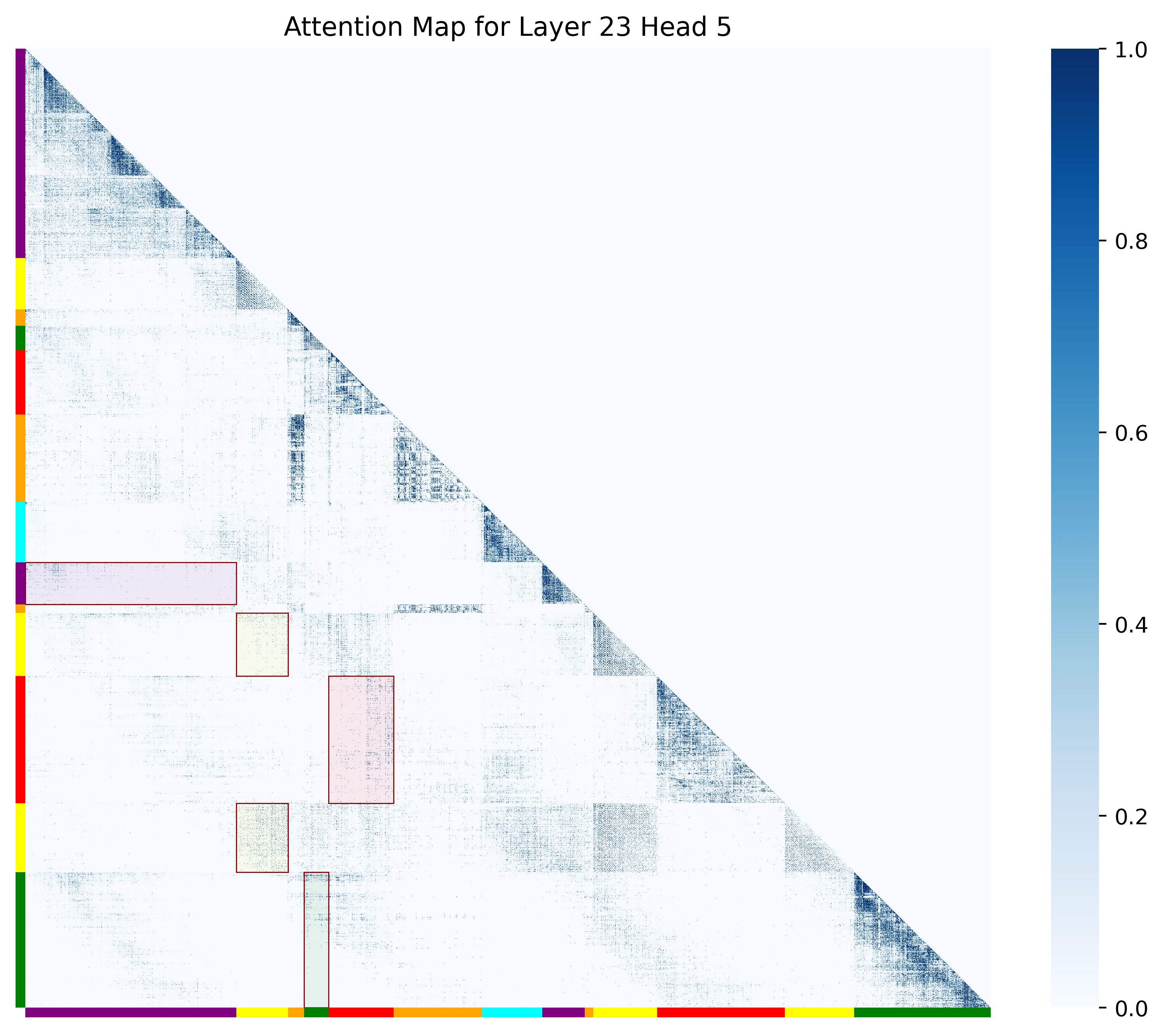}
        \caption{Qwen2-base 7B}
        \label{fig:attn1}
    \end{subfigure}
    \hfill
    \begin{subfigure}[b]{0.49\textwidth}
        \centering
        \includegraphics[width=\textwidth]{figs/attn-denoising-min.png}
        \caption{\modelname-base 7B}
        \label{fig:attn2}
    \end{subfigure}
    \vfill
    \begin{subfigure}[b]{0.49\textwidth}
        \centering
        \includegraphics[width=\textwidth]{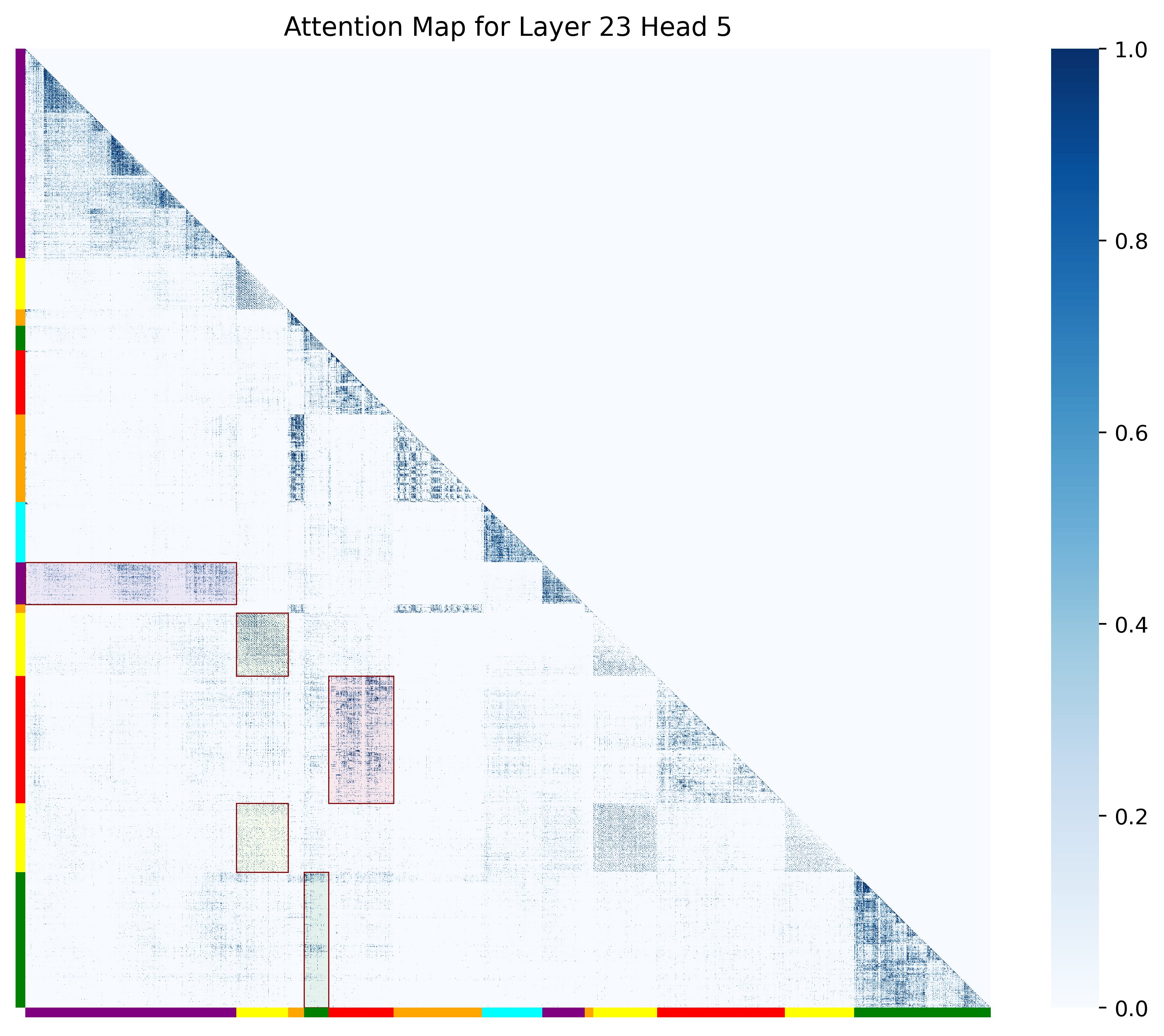}
        \caption{Qwen2-ABF-base 7B}
        \label{fig:attn3}
    \end{subfigure}
    \hfill
    \begin{subfigure}[b]{0.49\textwidth}
        \centering
        \includegraphics[width=\textwidth]{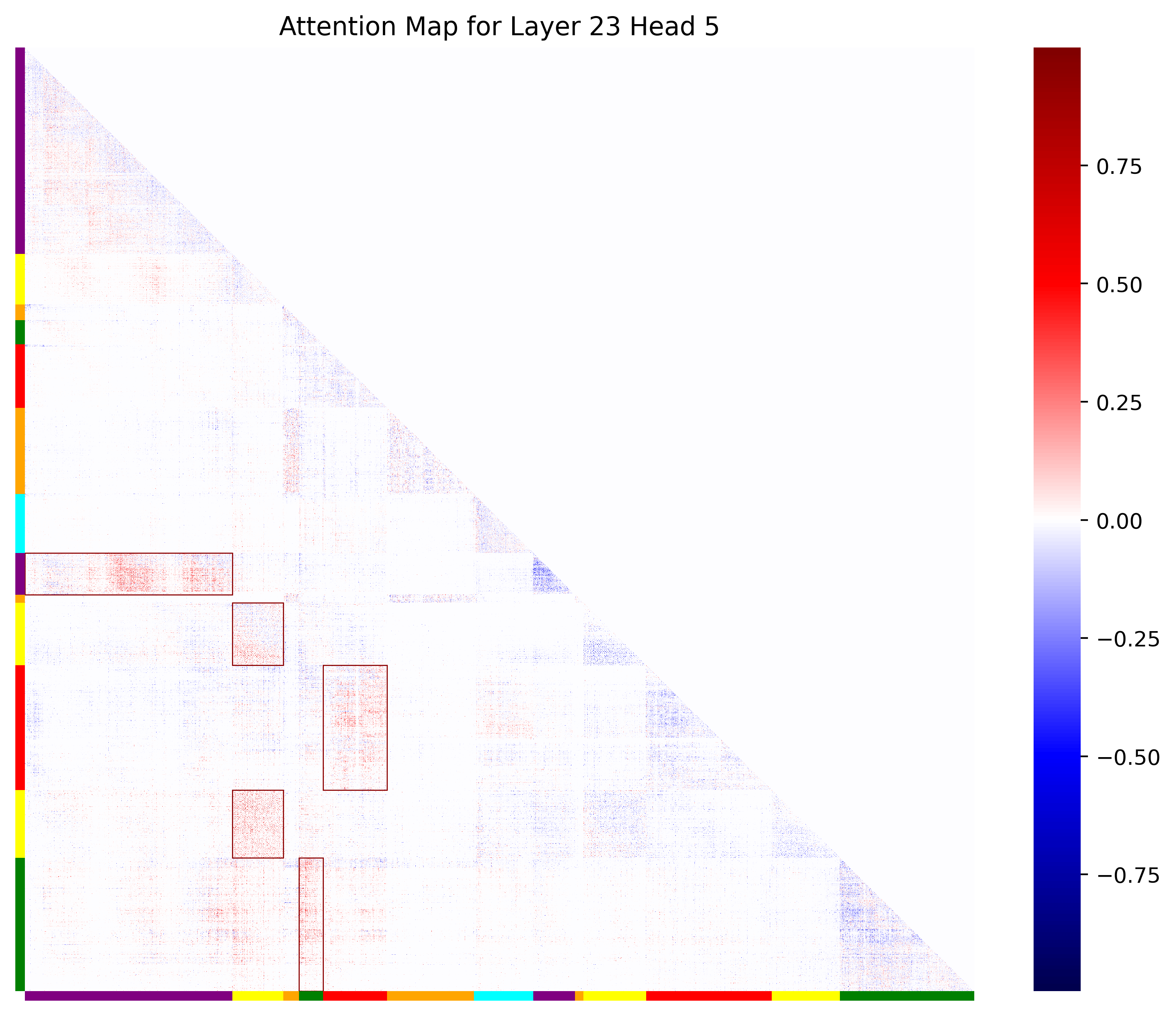}
        \caption{Qwen2-ABF-base v.s. \modelname-base }
        \label{fig:attn4}
    \end{subfigure}
    \caption{Attention Visualization}
    \label{fig:attn_combined}
\end{figure*}

Visualizing attention at 128k lengths presents some challenges. This is because at a length of 128k, with 28 layers and each layer having 28 attention heads, the attention scores would require $2 \times 28 \times 28 \times 128k \times 128k = 25\text{TB}$ (bf16) of memory. Therefore, during the forward pass, we only saved the $Q$ and $K$ for each layer to the disk. We then computed the attention score for each head of each layer offline, with each computation requiring only $2 \times 128k \times 128k = 32\text{GB}$ of memory. Due to the vast number of data points in the attention maps, we performed a pooling operation before plotting, retaining only the highest attention score within each 16x16 block. To emphasize the attention distribution, we multiplied each attention score by 100 and clipped the values to range between 0 and 1, resulting in an 8k x 8k attention map.

\subsection{Selection of Layers and Attention Heads}
Since most heads focus more on local attention, we needed to identify the layers and heads that represent long-range attention more effectively. We computed the sum of attention scores for distances greater than 1000 for each layer and attention head. Across multiple model calculations, we found that the head with the highest sum was the 5th head of the 23rd layer. Thus, we used the attention score of the 5th head of the 23rd layer for plotting.

\subsection{Document Splitting}
We selected six documents and concatenated them, resulting in a total of 147,917 tokens. We truncated any part exceeding 128k tokens. Each document was randomly split into three pieces, making a total of 18 chunks. Since the parts exceeding 128k tokens were truncated, only the first 16 chunks were used for computation. The coordinate axes in the plots display only 13 different slices because some slices from the same document remain adjacent even after shuffling.

\subsection{Explanation of Figures}
(a) The Qwen2-base 7B model is the original open-source model. During plotting, the support of DCA+YaRN in vLLM and HuggingFace caused out-of-memory (OOM) issues, so we did not include the YaRN+DCA strategy in the plot. It can be observed that for content beyond 32k tokens, the model shows very little attention score, indicating that the original model does not have the capability beyond 32k tokens.

(b) The Qwen2-ABF-base 7B is a model trained with the ABF strategy on 20B tokens. The ABF-trained baseline can accurately locate information within the same document.

(c) The Qwen2-UtK-base 7B is a model trained with the UtK strategy on 20B tokens. The plot shows that the UtK-trained model also accurately locates information within the same document.

(d) To compare the Qwen2-ABF-base 7B and Qwen2-UtK-base 7B models, we subtracted one attention score from the other and plotted the difference in figure (d). Red indicates higher attention scores for Qwen2-UtK-base, while blue indicates higher scores for Qwen2-ABF-base. The comparison reveals that the model trained with UtK shows more attention on long-range dependencies within the same document, thereby reducing the loss of long-range information.

\section{UtK Algorithm}
\label{sec:utk_algorithm}
Suppose $n$ documents represent a sampled set of training data of length $l$ (e.g., 128k), the $i$th document is represented as $\mathcal{D}_i$, which contains $\mathcal{L}_i$ tokens. $\sum_{i=1}^{n}{\mathcal{L}_i} >= l$.

UtK rearranges the training data in the following procedures: 
\begin{enumerate}
\item For each document $\mathcal{D}_i$ which $\mathcal{L}_i >= min\_split$, we split it into $h_i$ chunks, ${\mathcal{D}_i^1}$ to ${\mathcal{D}_i^{h_i}}$, $h_i \sim \mathcal{P}$, $\mathcal{P}$ is a custom discrete distribution, $2*(h_i-1)$ split points are randomly chosen from $(0, \mathcal{L}_i)$.
\item Prepend chunk label $CL_i^j$ for each chunk. Chunk labels are randomly generated characters, and are treated as normal words when doing tokenization. Each chunk label is surrounded by special tokens <CL> and </CL>.
\item For $\mathcal{D}_i^j$ with $j>1$, prepend head knot token <$h_j$>
\item For $\mathcal{D}_i^j$ with $j<h_i$, add tail knot token <$t_j$> at the end of this chunk.
\item Shuffle all $\mathcal{D}_i^j$s, when \textbf{PreserveOrder} constraint is enabled, we adjust the position of chunks of the same documents to preserve the order within each document.
\item Add untie solution at the last chunk of each document, <S> $CL_i^1$ <s> $CL_i^2$ <s> ... <s> $CL_i^{h_i}$ </S>.

\end{enumerate}
See Algorithm~\autoref{alg:utk-algo} for pseudo code implementation of UtK.

\begin{algorithm*}
\caption{UtK algorithm}\label{alg:utk-algo}
\begin{algorithmic}[1]
\Require $n > 0$
\Ensure $\sum{doc_i} > \textit{seq\_len}$
\Statex
\Procedure{BuildUtK}{docs}
\For {$i\gets 1, n$}
\Comment{Randomly split $doc_i$ into $h$ parts}
\State $s \gets []$
\State $parts \gets [doc_i]$
\If {$length(doc) \geq \textit{min\_split\_len}$}
\State {$h \gets \textit{random.choice([1..$max_h$, 1], $\mathcal{P}$])}$}
\Comment{Number of hops}
\State {$s \gets \textit{random.choice([1..length(} doc_i\textit{)), h)}$}
\Comment{Split position}
\State $s.sort()$
\EndIf
\State $parts_i \gets [doc_i[:s_1], doc_i[s_1:s2], ..., doc_i[s_{h-1}:]]$
\For {$j\gets 1, h$}
\Comment{Add Knot tokens before/after each doc}
\If {$j > 1$}
\State $parts_{ij} \gets [\text{"<}h_j\text{>"}] + parts_{ij}$
\EndIf
\If {$j < h$}
\State $parts_{ij} \gets parts_{ij} + [\text{"<}t_j\text{>"}]$
\EndIf
\State $parts_{ij} \gets \text{<ID>} + rand\_id_i + \text{</ID>} + parts_{ij}$
\Comment{Add random id}
\If {$j = h$}
\Comment{Add untie solution}
\State $parts_{ij} \gets parts_{ij} + \text{<ID>} + rand\_id_1 + ... + rand\_id_h + \text{</ID>}$
\EndIf
\EndFor
\EndFor
\State $total\_parts \gets \sum{length(parts_i)}$
\State $all\_indices \gets random.permutation(total\_parts)$
\State $start \gets 0$
\State $results \gets \text{list of size }total\_parts$
\For {$i\gets 1, n$}
\Comment{Gather parts of docs into a full sequence}
\State $this\_part\_indices \gets all\_indices[start:start + length(parts_i)]$
\State $this\_part\_indices.sort()$
\For {$j\gets 1, length(parts\_i)$}
\State $idx \gets this\_part\_indices[j]$
\State $results_{idx} \gets parts_{ij}$
\EndFor
\State $start \gets starts + length(parts_i)$
\EndFor
\Return results
\EndProcedure
\end{algorithmic}
\end{algorithm*}

\section{Additional Results}
\label{app:sec:results}


\begin{table*}[]
\centering
\setlength{\tabcolsep}{4pt} 
\begin{tabular}{lccccccc}
\toprule
Model & 4K & 8K & 16K & 32K & 64K & 128K \\
\midrule
Llama3.1-UtK-base (8B) & 94.64 & 92.19 & 91.73 & 88.83 & 83.60 & 73.79 \\
Llama3.1-base (70B) & 95.78 & 94.54 & 93.04 & 91.66 & 84.64 & 66.02 \\
Llama3.1-base (8B) & 94.35 & 92.06 & 92.31 & 90.17 & 80.40 & 66.10 \\
Qwen2-ABF (7B) &  99.78	& 98.53  & 82.46 & 78.94 & 75.21 & 65.91 \\
Qwen2-AttnMask (7B) & 90.57 & 84.9 & 82.74 & 80.38 & 75.59 & 71.97 \\
Qwen2-CIP (7B) & 90.5 & 85.38 & 82.21 & 80.50 & 76.26 & 71.04 \\
Qwen2-CT-base (7B) & 92.82 & 85.79 & 83.12 & 78.16 & 73.64 & 54.75 \\
Qwen2-Synthetic (7B) &  99.72 & 99.16  & 85.55 & 83.21 & 80.45 & 72.68 \\
Qwen2-Upsampling (7B) & 91.75 & 87.32 & 82.76 & 80.69 & 76.38 & 67.41 \\
Qwen2-UtK-base (72B) & 95.0 & 93.78 & 94.67 & 93.26 & 90.57 & 84.45 \\
Qwen2-UtK-base (7B) & 90.59 & 85.01 & 82.01 & 80.50 & 79.20 & 75.03 \\
Qwen2-base (72B) & 96.91 & 95.69 & 94.53 & 93.31 & 85.87 & 78.00 \\
Qwen2-base (7B) & 90.81 & 84.78 & 82.33 & 81.05 & 73.16 & 65.22 \\
\bottomrule
\end{tabular}
\caption{Performance of the reported base models across length 4K to 128K  by averaging 13 task scores of RULER.}
\end{table*}

\begin{table*}[]
\centering
\setlength{\tabcolsep}{3pt}
\begin{tabular}{lcccccc|cccccccc}
\toprule
\multirow{2}{*}{Model} & \multicolumn{6}{c|}{NIAH} & \multicolumn{6}{c}{VT} \\
 & 4K & 8K & 16K & 32K & 64K & 128K & 4K & 8K & 16K & 32K & 64K & 128K \\
\midrule
Llama3.1-UtK-base (8B) & 99.88 & 99.88 & 99.59 & 98.19 & 97.34 & 88.25 & 93.6 & 90.6 & 91.8 & 94.4 & 89.2 & 65.0 \\
Llama3.1-base (70B) & 100.0 & 99.62 & 99.59 & 97.56 & 95.09 & 74.88 & 94.4 & 94.0 & 94.8 & 85.4 & 83.6 & 75.0 \\
Llama3.1-base (8B) & 99.88 & 100.0 & 99.72 & 99.03 & 94.66 & 81.53 & 95.8 & 92.4 & 94.6 & 92.4 & 88.8 & 31.0 \\
Qwen2-ABF (7B) &  99.78	& 98.53	  & 98.16 & 95.53 & 93.72 & 83.06 &  78.4 &	80.2  & 71.8 & 66.2 & 65.6 & 71.6 \\
Qwen2-AttnMask (7B) & 98.38 & 98.09 & 97.38 & 96.69 & 92.09 & 86.34 & 72.4 & 60.4 & 58.0 & 48.4 & 46.6 & 76.4 \\
Qwen2-CIP (7B) & 99.38 & 99.31 & 98.22 & 95.97 & 93.50 & 86.12 & 63.8 & 65.4 & 65.0 & 69.0 & 72.4 & 90.4 \\
Qwen2-CT-base (7B) & 99.84 & 99.09 & 98.50 & 94.88 & 93.12 & 66.72 & 91.6 & 71.2 & 63.4 & 61.4 & 66.2 & 68.4 \\
Qwen2-Synthetic (7B) &  99.72 & 99.16  & 98.75 & 96.91 & 96.09 & 89.97 &  98.4 & 99.6  & 93.8 & 96.0 & 96.4 & 92.4 \\
Qwen2-Upsampling (7B) & 99.47 & 99.12 & 98.66 & 97.44 & 95.78 & 87.28 & 71.4 & 70.8 & 62.4 & 60.8 & 57.0 & 61.6 \\
Qwen2-UtK-base (72B) & 99.34 & 98.69 & 99.78 & 98.69 & 98.59 & 96.59 & 89.6 & 92.0 & 95.0 & 98.4 & 98.6 & 97.6 \\
Qwen2-UtK-base (7B) & 99.78 & 99.0 & 98.25 & 97.38 & 95.19 & 90.25 & 55.8 & 57.8 & 60.2 & 63.4 & 80.2 & 97.6 \\
Qwen2-base (72B) & 100.0 & 99.69 & 99.50 & 98.66 & 91.50 & 84.81 & 96.2 & 97.8 & 98.0 & 98.6 & 95.6 & 94.2 \\
Qwen2-base (7B) & 99.62 & 98.94 & 97.97 & 95.22 & 86.78 & 78.31 & 47.6 & 53.6 & 48.2 & 76.0 & 69.0 & 62.0 \\
\bottomrule
\end{tabular}
\caption{Performance of RULER's Retrieval (NIAH) and Multi-hop Tracing (VT) tasks across context lengths from 4K to 128K, averaged over 8 task scores for NIAH and 1 task score for VT.}
\end{table*}

\begin{table*}[]
\centering
\setlength{\tabcolsep}{3pt}
\begin{tabular}{lcccccc|cccccccc}
\toprule
\multirow{2}{*}{Model} & \multicolumn{6}{c|}{CWE+FWE} & \multicolumn{6}{c}{QA} \\
 & 4K & 8K & 16K & 32K & 64K & 128K & 4K & 8K & 16K & 32K & 64K & 128K \\
\midrule
Llama3.1-UtK-base (8B) & 95.84 & 90.44 & 90.00 & 73.44 & 49.95 & 43.14 & 73.0 & 64.0 & 62.0 & 64.0 & 59.5 & 51.0 \\
Llama3.1-base (70B) & 99.84 & 97.5 & 97.98 & 98.38 & 74.52 & 43.62 & 75.5 & 71.5 & 61.0 & 64.5 & 53.5 & 48.5 \\
Llama3.1-base (8B) & 96.36 & 91.72 & 92.85 & 83.76 & 47.05 & 38.56 & 69.5 & 60.5 & 61.0 & 60.0 & 52.5 & 49.5 \\
Qwen2-ABF (7B) &  89.65	& 68.48  & 52.95 & 49.89 & 38.71 & 23.35 &  69.0 & 59.0 & 54.5 & 48.0 & 42.5 & 37.0 \\
Qwen2-AttnMask (7B) & 85.98 & 61.3 & 53.28 & 50.98 & 41.68 & 43.26 & 73.0 & 68.0 & 66.0 & 60.5 & 58.0 & 41.0 \\
Qwen2-CIP (7B) & 86.3 & 62.05 & 51.50 & 50.86 & 43.98 & 33.55 & 72.5 & 63.0 & 57.5 & 54.0 & 41.5 & 38.5 \\
Qwen2-CT-base (7B) & 90.18 & 67.68 & 56.58 & 47.84 & 33.58 & 16.31 & 68.0 & 58.0 & 58.0 & 50.0 & 39.5 & 38.5 \\
Qwen2-Synthetic (7B) & 94.82 & 79.05  & 57.16 & 54.74 & 44.85 & 31.82 & 64.5 & 60.0 & 57.0 & 50.5 & 45.5 & 34.5 \\
Qwen2-Upsampling (7B) & 90.8 & 75.2 & 55.58 & 53.35 & 39.35 & 21.74 & 72.0 & 60.5 & 56.5 & 51.0 & 45.5 & 36.5 \\
Qwen2-UtK-base (72B) & 99.34 & 97.78 & 96.25 & 95.20 & 77.54 & 58.25 & 76.0 & 71.0 & 72.5 & 67.0 & 67.5 & 55.5 \\
Qwen2-UtK-base (7B) & 88.84 & 65.68 & 53.98 & 50.56 & 44.94 & 29.92 & 73.0 & 62.0 & 56.0 & 51.5 & 49.0 & 48.0 \\
Qwen2-base (72B) & 99.84 & 97.85 & 94.42 & 95.58 & 80.37 & 70.16 & 82.0 & 76.5 & 73.0 & 67.0 & 64.0 & 50.5 \\
Qwen2-base (7B) & 94.46 & 66.54 & 58.66 & 55.96 & 48.90 & 40.71 & 73.5 & 62.0 & 60.5 & 52.0 & 45.0 & 39.0 \\
\bottomrule
\end{tabular}
\caption{Performance of RULER's aggregation (CWE+FWE) and question answering (QA) tasks across context lengths from 4K to 128K, averaged over 2 task scores for CWE+FWE and 2 task scores for QA.}
\end{table*}


\begin{table*}[]
\centering
\setlength{\tabcolsep}{3pt} 
\begin{tabular}{lcccccccccc}
\toprule
 Models & cmrc & dureader & \makecell{hotpot\\wikiqa} & lic & \makecell{loogle\\CR} & \makecell{loogle\\MIR} & \makecell{loogle\\SD} & \makecell{mfqa\\en} & \makecell{mfqa\\zh} & \makecell{Avg.\\F1} \\
\midrule
Llama3.1-base (8B) & 39.15 & 13.55 & 22.60 & 16.77 & 14.93 & 13.31 & 45.25 & 20.95 & 28.59 & 23.90 \\
Qwen2-base (7B) & 48.88 & 15.76 & 22.67 & 15.36 & 11.34 & 8.68 & 40.93 & 26.22 & 25.60 & 23.94 \\
\rowcolor{gray!30} Llama3.1-UtK-base (8B) & 47.99 & 14.42 & 24.63 & 22.40 & 14.74 & 14.48 & 48.44 & 27.65 & 27.30 & 26.89 \\
\rowcolor{gray!30} \modelname-base (7B) & \bf 55.85 & 18.88 & 25.94 & \bf 24.42 & 15.77 & 14.34 & 43.96 & \bf 32.17 & 26.98 & 28.70 \\
\midrule
LLama3.1-base (70B) & 31.82 & 13.46 & 21.08 & 17
.08 & 18.92 & 13.02 & 44.01 & 20.47 & 27.76 & 23.07 \\
Qwen2-base (72B) & 44.64 & 20.76 & 24.68 & 18.68 & 16.37 & 16.62 & 48.78 & 26.17 & 29.91 & 27.40 \\
\rowcolor{gray!30} \modelname-base (72B) & 55.46 & \bf 21.04 & \bf 35.18 & 21.08 & \bf 19.03 & \bf 16.94 & \bf 56.96 & 29.13 & \bf 34.12 & \bf 32.10 \\
\bottomrule
\end{tabular}
\caption{Performance of LV-Eval at 128K context length, averaged across 9 question answering task scores.}
\end{table*}

\begin{table*}[]
\centering
\setlength{\tabcolsep}{3pt} 
\begin{tabular}{lcccccccccc}
\toprule
 Models & cmrc & dureader & \makecell{hotpot\\wikiqa} & lic & \makecell{loogle\\CR} & \makecell{loogle\\MIR} & \makecell{loogle\\SD} & \makecell{mfqa\\en} & \makecell{mfqa\\zh} & \makecell{Avg.\\F1} \\
\midrule
Llama3.1-base (8B) & 53.79 & 14.52 & 20.23 & 23.05 & 17.83 & 14.09 & \bf 59.03 & 28.61 & 31.30 & 29.16 \\
Qwen2-base (7B) & 58.85 & 17.90 & 29.79 & 21.32 & 13.16 & 15.86 & 54.17 & 26.54 & 31.32 & 29.88 \\
\rowcolor{gray!30} Llama3.1-UtK-base (8B) & \bf 61.27 & 15.55 & 21.81 & \bf 24.50 & 18.31 & 13.10 & 56.82 & 28.52 & 26.82 & 29.63 \\
\rowcolor{gray!30} \modelname-base (7B)  & 55.35 & 17.10 & 32.24 & 23.18 & 13.84 & 14.43 & 48.86 & 27.69 & 25.03 & 28.64 \\
\midrule
LLama3.1-base (70B) & 53.07 & 14.83 & 29.67 & 19.35 & \bf 22.84 & 18.00 & 55.02 & \bf 29.12 & 31.54 & 30.38 \\
Qwen2-base (72B)  & 57.58 & 20.86 & \bf 32.48 & 21.06 & 21.46 & 18.54 & 58.52 & 25.08 & \bf 35.71 & 32.37 \\
\rowcolor{gray!30} \modelname-base (72B) & 58.09 & \bf 22.54 & 31.97 & 22.49 & 19.69 & \bf 19.33 & 58.37 & 26.17 & 31.52 & \bf 32.24 \\
\bottomrule
\end{tabular}
\caption{Performance of LV-Eval at 32K context length, averaged across 9 question answering task scores.}
\end{table*}